\documentclass[letterpaper, 10 pt, conference]{ieeeconf}  

\IEEEoverridecommandlockouts                              

\usepackage{graphics} 
\usepackage{epsfig} 
\usepackage{mathptmx} 
\usepackage{times} 
\usepackage{amsmath} 
\usepackage{amssymb}  
\usepackage{xcolor}
\usepackage{multirow}
\usepackage{adjustbox}
\usepackage{booktabs}
\usepackage{subcaption}
\usepackage{caption}
\usepackage{cite}
\usepackage{array}
\usepackage{siunitx}
\usepackage{symbols}
\usepackage{color,soul}
\usepackage{float}
\usepackage{stfloats}
\usepackage{graphicx}
\usepackage{hyperref}
\usepackage{tabularx}
\usepackage{subcaption}
\usepackage{newtxtext, newtxmath}

\title{\LARGE \bf
THREAD: Trajectory Planning for Hybrid Rigid-Soft Manipulators\\ with Environment-Aware Diffusion

}

\author{Shivani Kamtikar$^{1}$, Pranav Asthana$^{2}$, Naveen Kumar Uppalapati$^{1}$, Girish Krishnan$^{1}$, Girish Chowdhary$^{1}$ \\ $^{1}$University of Illinois Urbana-Champaign $^{2}$University of Maryland College Park
}

\begin{document}

\maketitle
\thispagestyle{empty}
\pagestyle{empty}

\begin{abstract}

Manipulation in confined environments, such as threading a manipulator through narrow apertures, remains a fundamental challenge, especially for conventional rigid robots. Hybrid rigid-soft manipulators offer promise but face two compounding planning challenges: backbone shapes feasible in free space become infeasible under environmental contact, and planning rigid and soft segments independently ignores their kinematic coupling. We present~\ourmethod{}, the first diffusion-based trajectory planner for hybrid manipulation, learning a generative prior over physically realizable \emph{backbone trajectories} conditioned on local environment geometry, with physics-inspired losses encoding curvature, smoothness, and collision constraints jointly across both segments. Trained in simulation, \ourmethod{} achieves 92.4\% task success with 5$\times$ fewer collisions than the strongest baseline. We show cross-embodiment real-world transfer with minimal online updates, successfully threading through apertures as small as 1.3$\times$ the soft segment diameter. \url{https://robot-thread.github.io}

\end{abstract}

\section{Introduction}

\noindent Robots deployed in disaster response, search-and-rescue, infrastructure inspection, and subterranean exploration, must navigate debris, crevices, and narrow voids to reach targets inaccessible to conventional rigid manipulators~\cite{hawkes2017soft}. Hybrid rigid-soft manipulators have emerged as a promising solution to this challenge by combining the reach and stability of rigid arms with the compliance of soft continuum arms (SCAs), enabling systems to extend into tight spaces, and thread through constrained passages~\cite{uppalapati2020valens, xu2024hybrid, koe2025learning}. However, this hybrid architecture introduces a planning challenge that existing methods are ill-equipped to handle. Given only a target end-effector pose, several backbone shapes are kinematically consistent in free space, yet the vast majority are physically infeasible in complex environments – colliding with environment geometry and violating strain limits under contact. Further, planning rigid and soft segments independently ignores their kinematic coupling -- a poorly placed rigid arm may force the soft segment into configurations that violate strain limits or render the task infeasible altogether. \looseness=-1

We argue that the right inductive bias for this problem is a \emph{generative prior over physically realizable backbone shapes}, conditioned on the environment geometry the robot currently faces. Rather than searching for a single feasible trajectory, this prior represents the full distribution of valid solutions, enabling diverse candidate sampling, explicit coverage of the feasibility manifold, and graceful degradation as geometric constraints tighten. Diffusion models are a natural instantiation of this prior as they are expressive generative models capable of representing complex multi-modal distributions that have demonstrated strong planning performance in rigid manipulation~\cite{chi2025diffusion, wolf2025diffusion}. Yet they have never been applied to the fundamentally different domain of continuum manipulators, where the action space is a continuous backbone shape rather than a discrete joint configuration.\looseness=-1

\begin{figure}[t!]
    \centering
    \includegraphics[width=\linewidth]{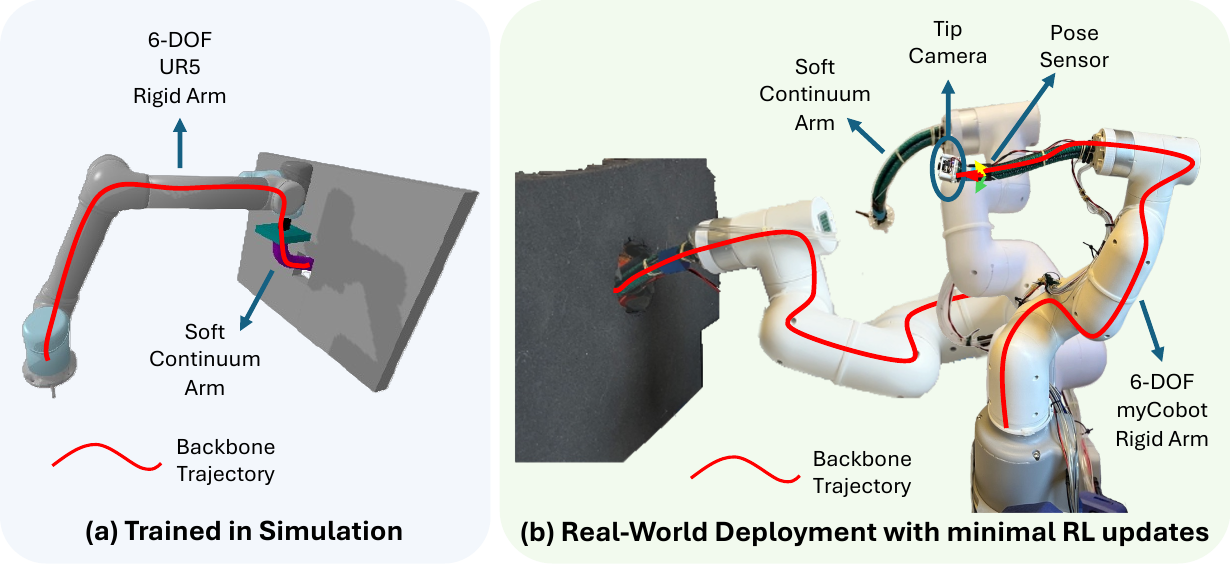}
    \caption{\textbf{Hybrid system setup.} Our system is trained in simulation~\cite{kasaeisoftmanisim} (a) using a two-segment soft continuum arm (SCA) attached to a 6-DOF UR5 rigid arm. It is deployed directly to our real-world setup (b) a single segment SCA attached to a 6-DOF myCobot 280 rigid arm, with minimal online RL adaptation. Planning in the backbone shape space enables transfer across different rigid-soft embodiments without retraining the planner.}
    \label{fig:system_fig}
    \vspace{-6mm}
\end{figure}

\noindent We present~\ourmethod{}, a diffusion-based trajectory planner for threading hybrid manipulators through constrained apertures. ~\ourmethod{} makes the following contributions.

\noindent1) \textbf{Unified rigid-soft backbone representation.} We encode the full kinematic chain as a backbone shape, enabling joint trajectory optimization over both rigid and soft subsystems.

\noindent2) \textbf{Diffusion planner in backbone shape space.} We introduce the first diffusion planner for hybrid manipulators, extending diffusion trajectory planning from rigid joint spaces to the continuous backbone shape domain.

\noindent3) \textbf{Inference-time collision avoidance with online reinforcement learning (RL) refinement.} We incorporate inference-time gradient-based geometric guidance into the diffusion sampling process to enforce collision-free trajectories, complemented by a residual RL policy that issues fine corrective updates during execution. 

\noindent4) \textbf{Sample-efficient cross-embodied real-world transfer.} We demonstrate that planning over backbone shapes enables direct transfer to a different rigid-soft embodiment in the real-world, requiring only minimal online RL adaptation without retraining the diffusion planner.

\section{Related Work}
\noindent\textbf{Diffusion Models for Robotic Manipulation.} Diffusion models have emerged as a dominant paradigm for learning robot behavior from demonstrations. Janner~\textit{et. al.}~\cite{janner2022planning} first demonstrated that classifier-guided sampling and inpainting could be reinterpreted as coherent planning strategies by iteratively denoising full state-action trajectories.
Diffusion Policy~\cite{chi2025diffusion} extended this to visuomotor control, showing that diffusion policies can represent complex multimodal action distributions, properties especially valuable in constrained environments. Subsequent work has incorporated diffusion as motion planning priors~\cite{carvalho2023motion}, explored inference-time constraint satisfaction via classifier guidance~\cite{wolf2025diffusion}, and integrated diffusion with RL for closed-loop reward-maximizing behavior~\cite{wolf2025diffusion}. Diffusion has also been applied to joint-space planning for cable-driven manipulators~\cite{hu2026fault} and soft robot co-design~\cite{wang2023diffusebot}. In contrast, \ourmethod{} is the first diffusion-based trajectory planner for hybrid rigid-soft manipulators, jointly optimizing over coupled subsystems in a continuous backbone shape space, where tight geometric constraints make collision-free planning substantially more challenging than in rigid single-robot settings.

\begin{figure*}[htp!]
\centering
\smallskip
  \includegraphics[width=1\textwidth]{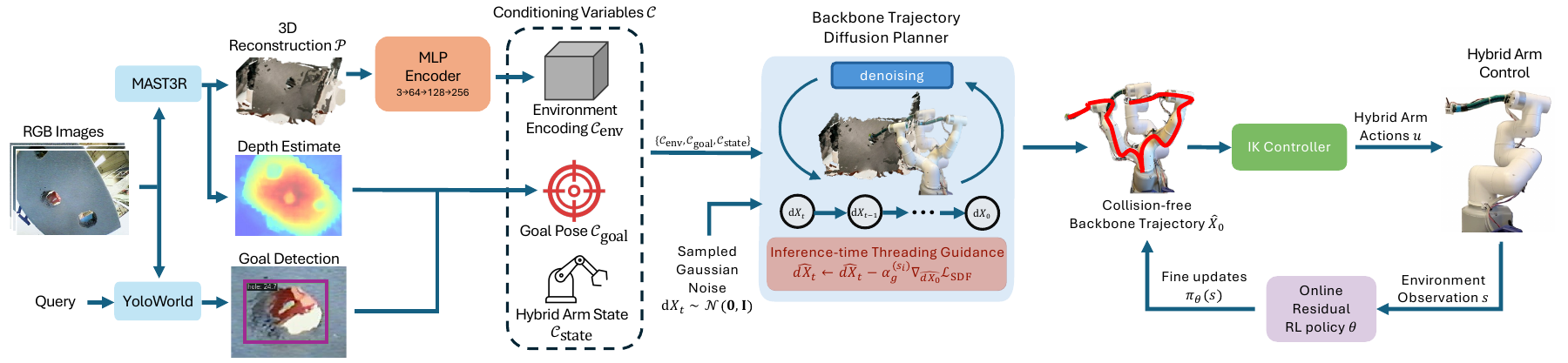}
  \caption{\textbf{Pipeline of \ourmethod{}.} Given a set of unposed RGB images from a tip-mounted camera, scene geometry is reconstructed using MASt3R~\cite{leroy2024grounding} and encoded by a learned point encoder to produce environment encoding $\mathcal{C}_{\text{env}}$. The goal pose $\mathcal{C}_{\text{goal}}$ is detected using YoloWorld~\cite{cheng2024yolo} and lifted to 3D using MASt3R depth. A backbone trajectory diffusion planner generates collision-free trajectories $\hat{X}_0$ conditioned on the environment encoding $\mathcal{C}_{\text{env}}$, goal pose $\mathcal{C}_{\text{goal}}$, and hybrid arm state $\mathcal{C}_{\text{state}}$, with inference-time threading guidance to enforce collision avoidance. A learned inverse kinematics (IK) controller converts backbone shape to hybrid arm control commands $u$. Finally, an online residual RL policy $\theta$ issues fine corrective updates from environment observations $s$, compensating for modeling error and contact uncertainty.
  This framework enables threading a hybrid manipulator through narrow apertures with high success and low collision rates.
  }
  \label{fig:pipeline}
  \vspace{-7mm}
\end{figure*}

\noindent \textbf{Continuum Robot Planning in Confined Spaces.} With compliant dexterous geometries, continuum robots achieve reach through narrow and constrained environments, enabling otherwise unachievable tasks for medical, industrial, and service applications. Classical sampling-based planners~\cite{li2024s, russo2023continuum} do not generalize across environments and lack closed-loop replanning, while supervised learning~\cite{koe2025learning, kamtikar2022visual,kamtikar2026hyreach} and RL-based approaches~\cite{nazeer2024comparison, nazeer2024rl, chen2025hysteresis} are limited to end-effector control, lacking global backbone reasoning or explicit geometric collision awareness. In contrast, \ourmethod{} directly operates in the full backbone shape space via a learned diffusion prior, incorporates inference-time collision avoidance through geometry-guided sampling, and closes the loop with residual RL correction. \looseness=-1


\noindent \textbf{Backbone Representation and Physics-Informed Learning.} Classical backbone modeling relies on constant-curvature approximations~\cite{he2023modeling} and Cosserat rod formulations~\cite{shi2023position}, which are computationally expensive and must be re-solved per configuration, making them ill-suited for real-time planning. Shape-aware frameworks~\cite{bensch2024physics} have augmented such models with neural residuals for MPPI-based planning, but still require per-configuration solves.
~\ourmethod{} entirely sidesteps forward-model inversion by learning a generative prior over backbone trajectories, replacing per-configuration solve complexity with a one-time demonstration dataset to train the prior and a single point cloud reconstruction computed once per scene at inference.\looseness=-1



\section{Method}
\everypar{\looseness=-1}

\noindent We address the problem of threading a hybrid manipulator through a narrow hole in a wall, requiring perception, motion planning, actuation, and online error correction. Our system, Figure~\ref{fig:pipeline}, comprises of an initial 3D reconstruction and goal detection phase followed by three components operating in a closed loop: (1) a diffusion trajectory planner that generates a backbone shape trajectory conditioned on the arm state, scene geometry and goal pose; (2) an inverse kinematics (IK) model that translates each planned shape into hybrid actuation commands; and (3) a residual RL policy that issues small corrective actions to close the gap between planned and executed trajectories. The planner is re-invoked in a receding-horizon replanning loop every 32 steps to re-anchor the plan to the current robot state.

\subsection{Scene Reconstruction and Goal Detection}
\noindent In simulation, the robot uses an in-hand RGB-D camera to construct a point cloud representation of the local environment, transformed into the robot base frame. The goal pose (hole center) is obtained via known geometry. In real-world, the hybrid robot performs a one-shot scene capture by exploring a small region around its home position, collecting RGB images from multiple views. The images are input to MASt3R~\cite{leroy2024grounding} to estimate metric depth and reconstruct the scene. The goal aperture is localized using YOLO-World~\cite{cheng2024yolo}, which uses natural-language queries enabling open-world flexibility and avoids dependence on a fixed set of objects. The detected hole locations are fused with the depth map to estimate 3D goal positions.

\subsection{Hybrid Backbone Representation}
\label{backbone-rep}
\noindent We represent the full kinematic chain -- a rigid manipulator with an SCA mounted at its end-effector (see Section~\ref{shape2act} for hardware details) -- as a single backbone of $K=64$ points is uniformly sampled along the robot from base to tip. 
A full demonstration episode is represented as a trajectory $\mathbf{X} \in \mathbb{R}^{H \times K \times 3}$, where $H=32$ frames are uniformly sampled in time from collected expert demonstrations.
At each frame $h$, the backbone shape $\mathbf{X}^h \in \mathbb{R}^{K \times 3}$ captures the spatial configuration of the full arm in the rigid arm base frame, and is equivalently expressed as segment-wise differences $\mathbf{dX}^h \in \mathbb{R}^{(K-1) \times 3}$, where $\mathbf{dX}^{h,k} = \mathbf{X}^{h, k+1} - \mathbf{X}^{h,k}$. 
A rigidity mask, parametrized by attachment index $k_{\text{attach}}$, divides the $K$ backbone points into rigid ($0{:}k_{\text{attach}}$) and soft ($k_{\text{attach}}{:}K-1$) segments.
While the rigid commands are recovered from $0{:}k_{\text{attach}}$ via IK, diffusing over the full backbone is not equivalent to diffusing over the end-effector plus SCA: the interior rigid points enforce that the generated end-effector pose remains realizable under the rigid arm's kinematics, which an end-effector-only formulation cannot guarantee. This unified representation enables a single model to reason over both rigid and deformable subsystems jointly, with physics-inspired loss terms enforcing appropriate geometric constraints in each region (Section~\ref{sec:diff_planner}).

\subsection{Diffusion Trajectory Planner}
\label{sec:diff_planner}
\noindent\textbf{Diffusion process.} We train a denoising diffusion probabilistic model (DDPM)~\cite{ho2020denoising} over backbone-delta trajectories $\mathbf{dX} \in \mathbb{R}^{H \times (K-1) \times 3}$ in simulation. The forward process corrupts a clean trajectory $\mathbf{dX}_0$ over $T=1000$ steps with the corrupted sample at timestep $t$ being distributed as:

\begin{equation}
    q(\mathbf{dX}_t | \mathbf{dX}_{t-1}) = \mathcal{N}(\mathbf{dX}_t; \sqrt{1-\beta_t}\mathbf{dX}_{t-1},\beta_t I)
\end{equation}
where $\beta_t$ follows a linear noise schedule with $\beta_t = \beta_1 + \frac{t-1}{T-1}(\beta_T-\beta_1)$ ($\beta_1 = 10^{-4}$, $\beta_T = 0.02$).
This can be efficiently sampled in closed form by the reparameterization trick as
\begin{equation}
    \mathbf{dX}_t = \sqrt{\bar\alpha_t}\,\mathbf{dX}_0 + \sqrt{1-\bar\alpha_t}\,\epsilon
\end{equation}
where $\bar\alpha_t = \prod_{i=1}^{t}(1-\beta_i)$ and $\epsilon \sim \mathcal{N}(\mathbf{0}, \mathbf{I})$.
The model is trained to predict the added noise $\epsilon$ from noisy input $\mathbf{dX}_t$, conditioning variables $\mathcal{C}$, and timestep $t$.
We apply z-score normalization to the delta representation $\mathbf{dX}_0$ before applying noise.

\noindent Unlike prior diffusion planners that encode scene context as a single global vector added to trajectory tokens~\cite{tevet2022human}, we adopt the cross-attention conditioning paradigm~\cite{chi2025diffusion} where each of the $H=32$ trajectory-time tokens attends to a sequence of heterogeneous conditioning tokens:

1) \textit{Point cloud tokens ($\mathcal{C}_{\text{env}}$ $\in \mathbb{R}^{128 \times 256}$)}: 
A subset of 128 points is uniformly sampled from the reconstructed point cloud $\mathcal{P}$ yielding 128 token vectors. An MLP ($3 \to 64 \to 128 \to 256$), trained end-to-end with the diffusion model, lifts each token in $\mathbb{R}^3$ to a $256$-dimensional feature, yielding $(128 \times 256)$ features that represent spatial structure across the scene.

2) \textit{Initial state token ($\mathcal{C}_{\text{state}}$ $\in \mathbb{R}^{1 \times 3(K-1)}$)}: The flattened initial backbone deltas $\mathbf{dX}^{h=0}$, encoding the robot's initial shape. \looseness=-1

3) \textit{Goal token ($\mathcal{C}_{\text{goal}}$ $\in \mathbb{R}^{1 \times 3}$)}: The 3D goal position, expressed as a displacement from the initial tip to encourage translation invariance.

\noindent Each conditioning token of native dimension $C_d$ is projected to the model 
dimension $d{=}512$ via a modality-specific learned linear projection 
$\mathbf{W}_m \in \mathbb{R}^{d \times C_d}$, and assigned a learnable type embedding 
$e_m \in \mathbb{R}^d$ to distinguish modalities, analogous to segment embeddings in BERT~\cite{devlin2019bert}.

\noindent\textbf{Training losses.} Following the DDPM reparameterization~\cite{ho2020denoising}, the model is trained to predict the noise $\epsilon$ added during the forward process. The primary loss is the frame-weighted noise prediction MSE:
\begin{equation}
    \mathcal{L}_{\text{MSE}} = \mathbb{E}_{h,k}\left[
    w_h \left\|\hat{\boldsymbol{\epsilon}}_\theta^{h,k} - \boldsymbol{\epsilon}^{h,k}\right\|^2_2
    \right]
\end{equation}

\noindent where $\hat{\boldsymbol{\epsilon}}_\theta^{h,k} \in \mathbb{R}^3$ is the model's predicted noise at frame $h$ and backbone segment $k$, taken from the full output 
$\hat{\boldsymbol{\epsilon}}_\theta(\mathbf{dX}_t, t, \mathcal{C}) \in \mathbb{R}^{H \times (K-1) \times 3}$,
$\mathcal{C}$ is the conditioning (point cloud, initial state, goal),
and $w_h$ increases linearly from $1$ to $3$ over the $H$ frames, 
up-weighting goal-reaching frames.

\noindent We additionally apply three geometric regularizers:

1) \textit{Curvature loss} $\mathbf{\mathcal{L}_c}$ matches the second-order spatial curvature of the prediction to that of the ground truth, encouraging physically plausible backbone shapes:
    \begin{equation}
        \mathcal{L}_c = \left|\|\nabla^2_k \hat{\mathbf{dX}}_0\|_2^2
         - \|\nabla^2_k \mathbf{dX}_0\|_2^2\right|
    \end{equation}
    
2) \textit{Temporal smoothness loss} $\mathbf{\mathcal{L}_t}$ penalizes acceleration in the trajectory (second-order temporal differences), producing smooth frame transitions:
    \begin{equation}
        \mathcal{L}_t = \|\nabla^2_h \hat{\mathbf{dX}}_0\|_2^2
        \label{eq:smoothness_loss}
    \end{equation}

3) \textit{Rigid straightness loss} $\mathbf{\mathcal{L}_r}$ penalizes curvature in the rigid portion $\mathcal{R} = \{k : k <k_{\text{attach}}\}$, enforcing that rigid links remain straight:
    \begin{equation}
        \mathcal{L}_r = \frac{1}{|\mathcal{R}|}\sum_{k \in \mathcal{R}}
        \|\nabla^2_k \hat{\mathbf{dX}}_0\|_2^2
    \end{equation}

The total loss is:
\begin{equation}
    \mathcal{L}_{\text{diffusion}} = 
    \lambda_{\text{MSE}}\mathcal{L}_{\text{MSE}} + 
    \lambda_c\mathcal{L}_c + 
    \lambda_t\mathcal{L}_t + 
    \lambda_r\mathcal{L}_r
\end{equation}
with $\lambda_{\text{MSE}}{=}1.0$, $\lambda_c{=}0.1$, $\lambda_t{=}0.05$, 
$\lambda_r{=}0.5$. Ablation study on diffusion losses and design is presented in Table~\ref{tab:ablation} and Figure~\ref{fig:ablation}. 

\noindent\textbf{Inference-time guidance.} At inference, we use DDIM~\cite{songdenoising} with $S$ sampling steps and quadratic timestep spacing, which concentrates denoising steps in the late, geometry-forming phase of the reverse process. We additionally guide diffusion sampling using a differentiable signed distance field (SDF) constructed from the observed scene point cloud.

\noindent Let $\Phi : \mathbb{R}^3 \rightarrow \mathbb{R}$ denote the SDF defined in the rigid-arm base frame, where $\Phi(\cdot) > 0$ indicates free space and $\Phi(\cdot) < 0$ indicates penetration into obstacles. 
The SDF for a backbone point $x$ is computed as the distance to the nearest point in the reconstructed point cloud $\mathcal{P}$, $\phi(x) = \min_{p\in\mathcal{P}}\|x-p\|_2$.
Given the predicted clean backbone $\hat{\mathbf{X}}_0 \in \mathbb{R}^{H \times K \times 3}$ reconstructed from $\hat{\mathbf{dX}}_0$, we compute an SDF-based collision loss over all backbone points:
\begin{equation}
\mathcal{L}_{\text{SDF}} 
= \lambda_{\text{col}} 
\sum_{h=1}^{H} 
\sum_{k=1}^{K}
\left[\text{ReLU}(-\Phi(\hat{\mathbf{X}}_0^{h,k}))\right]
\end{equation}

\noindent This penalizes backbone points that lie inside obstacles while leaving free-space configurations unaffected. 
During each DDIM sampling step $s_i$ in the guided portion ($s_i/S \geq 0.5$), we differentiate $\mathcal{L}_{\text{SDF}}$ with respect to the predicted clean trajectory $\hat{\mathbf{dX}}_0$ and update $\hat{\mathbf{dX}}_t$ before continuing denoising:

\begin{equation}
    \hat{\mathbf{dX}}_t \leftarrow \hat{\mathbf{dX}}_t - \alpha_g^{(s_i)} \nabla_{\hat{\mathbf{dX}}_0}\mathcal{L}_{\text{SDF}}
\end{equation}

\noindent where guidance strength increases linearly over the guided portion,

\begin{equation}
    \alpha_g^{(s_i)} = \lambda_{\text{SDF}} \cdot 
    \frac{s_i/S - f_{\text{start}}}{1 - f_{\text{start}}}
\end{equation}

\noindent with $\lambda_{\text{SDF}}{=}0.05$, $f_{\text{start}}{=}0.5$, and $S=700$ the total number of DDIM steps. The updated $\hat{\mathbf{dX}}_0$ is then used to compute the next noisy sample $\hat{\mathbf{dX}}_{t-1}$ via the standard DDIM step.

\subsection{Shape to Actuation Commands}
\label{shape2act}
\noindent Given a target backbone frame $\mathbf{X}^{h}$ from the planner, we obtain actuation commands $\mathbf{u}$ via platform specific mappings.

\noindent\textbf{Simulation.}
In simulation, $\mathbf{u} = [\mathbf{q}_s \in \mathbb{R}^6,\ \Delta \mathbf{p}_{\text{ee}} \in \mathbb{R}^3]$, corresponds to soft curvature parameters for the two-segment SCA and an end-effector position delta for the UR5, obtained via resolved-rate Jacobian IK~\cite{kasaeisoftmanisim}.

\noindent\textbf{Real-World.} On hardware, $\mathbf{u} = [\mathbf{b} \in \mathbb{R}^3,\ \Delta \mathbf{j} \in \mathbb{R}^6]$ corresponds to the three pneumatic actuation inputs for the single-segment SCA and six joint deltas for the rigid arm. Since analytic IK is unavailable for the real SCA, we learn an MLP to map observed backbone shapes to their corresponding actuation commands, followed by a closed-loop control law~\cite{kamtikar2022visual} to correct for inaccuracies in the learned model. The model is trained on 9,536 paired backbone-actuation samples collected from random exploration of the manipulator's actuation space. The SCA backbone shape is computed using a Piecewise Constant Curvature approximation~\cite{rao2021model, he2023modeling} discretized along its length, with rigid segment poses determined from joint angles and known geometry; intermediate points are interpolated to form a
$K=64$ point backbone.  \looseness=-1

\subsection{Residual RL with Receding-Horizon Replanning}

\noindent\textbf{Residual control formulation.}
While diffusion produces backbone trajectories with geometric structure, open-loop execution is insufficient under modeling error, compliance, and contact. We therefore train a residual Soft Actor-Critic (SAC) policy $\boldsymbol{\pi}_\theta$ that issues small corrective commands on top of the diffusion output $\hat{\mathbf{X}}_0$. The executed command at each frame $f$ is \looseness=-1
\begin{equation}
    \hat{\mathbf{X}}_0 \leftarrow \hat{\mathbf{X}}_0 + \boldsymbol{\pi}_{\theta}(s)
\end{equation}
where $\boldsymbol{\pi}_\theta(s)$ is the learned residual correction conditioned on observation $s$. Residual actions are clipped to $\pm 20$ mm to prevent the policy from overriding the geometric structure imposed by the planner.

\noindent\textbf{Observation space.}
The policy observes a compact 28-dimensional state comprising:
(1) tip tracking error ($\mathbb{R}^3$), 
(2) backbone tracking errors at 5 uniformly sampled points ($\mathbb{R}^{15}$),
(3) goal position relative to the tip ($\mathbb{R}^3$),
and 
(4) distance to the scene point cloud at the same 5 points ($\mathbb{R}^{5}$, clipped to $[0, 10\text{cm}]$). 
Threading is successful when the nearest scene point lies in the half-space opposite to the goal:

\begin{equation}
    (\mathbf{x}^\star -\mathbf{p}_{\text{tip}}) \cdot (\mathbf{p}_{\text{goal}} - \mathbf{p}_{\text{tip}}) < 0
\end{equation}

where $\mathbf{x}^\star = \arg\min_{\mathbf{x} \in \mathcal{P}} \|\mathbf{x} - \mathbf{p}_{\text{tip}}\|^2_2$ is the nearest point in the scene point cloud $\mathcal{P}$ and $\mathbf{p}_\text{tip}$ and $\mathbf{p}_\text{goal}$ are the tip and goal pose respectively.

\noindent\textbf{Reward.}
The residual policy is trained using a shaped reward that encourages
accurate tracking, collision avoidance, and smooth corrections using various rewards as follows:

1) \textit{Tracking reward} encourages the tip to follow the diffusion plan using a Gaussian radial basis function (RBF):
\begin{equation}
r_{\text{track}} = \exp\!\bigl(-\gamma\,\|\mathbf{p}_{\text{tip}} - \mathbf{p}_{\text{plan}}\|_2^2\bigr)
\end{equation}
where $\mathbf{p}_{\text{tip}}$ and $\mathbf{p}_{\text{plan}}$ are the actual and planned tip poses. RBF yields near-maximum reward when tracking error is small and decays sharply beyond $\sim$10 cm, controlled by the bandwidth parameter $\gamma=30$. This ensures the tracking signal remains well-scaled relative to the other reward terms without dominating the overall objective.

2) \textit{Collision penalty} penalizes backbone points that violate
a safety margin $r_{\text{safe}}$ around the nearest scene surface:

\begin{equation}
r_{\text{col}} = -\sum_{k \in \mathbb{S}} \mathrm{ReLU}\!\bigl(-|\Phi(\mathbf{X}^{h,k})| + r_{\text{safe}}\bigr),
\end{equation}

\noindent where $\mathbb{S}$ is a set of five uniformly sampled backbone indices, $r_{\text{safe}} = 15\,\text{mm}$ is a safety margin and $\mathbf{X}^{h,k}$ the $k^{\text{th}}$ backbone point at frame $h$.

3) \textit{Smoothness penalty} discourages large residual corrections:
\begin{equation}
r_{\text{smooth}} = -\|\boldsymbol{\pi}_\theta(s)\|_2^2
\end{equation}

4) \textit{Goal proximity} provides dense reward after threading the aperture,
gated by the threading condition $\mathbf{1}_{\text{thr}}$:
\begin{equation}
r_{\text{goal}} = \exp\!\bigl(-\gamma\,\|\mathbf{p}_{\text{tip}} - \mathbf{p}_{\text{goal}}\|_2^2\bigr)
\cdot \mathbf{1}_{\text{thr}}
\end{equation}

5) \textit{Terminal bonus} $r_{\text{terminal}} = +20$ is awarded when
$\|\mathbf{p}_{\text{tip}} - \mathbf{p}_{\text{goal}}\|_2 < 2\,\text{cm}$, ending the episode.

6) \textit{Time penalty} $c_t = 0.01$ per step encourages efficient
task completion.

\noindent The total reward at frame $f$ of the episode is:
\begin{equation}
\begin{split}
r =
&w_{\text{track}} \, r_{\text{track}}
+ w_{\text{col}} \, r_{\text{col}}
+ w_{\text{smooth}} \, r_{\text{smooth}}
+ w_{\text{goal}} \, r_{\text{goal}}\\
&+ r_{\text{terminal}}
- fc_t
\end{split}
\end{equation}
\noindent with $w_{\text{track}}{=}0.10$, $w_{\text{col}}{=}0.25$,
$w_{\text{smooth}}{=}0.05$, $w_{\text{goal}}{=}0.30$.

\begin{figure*}[htp!]
\centering
\smallskip
  \includegraphics[width=0.95\linewidth]{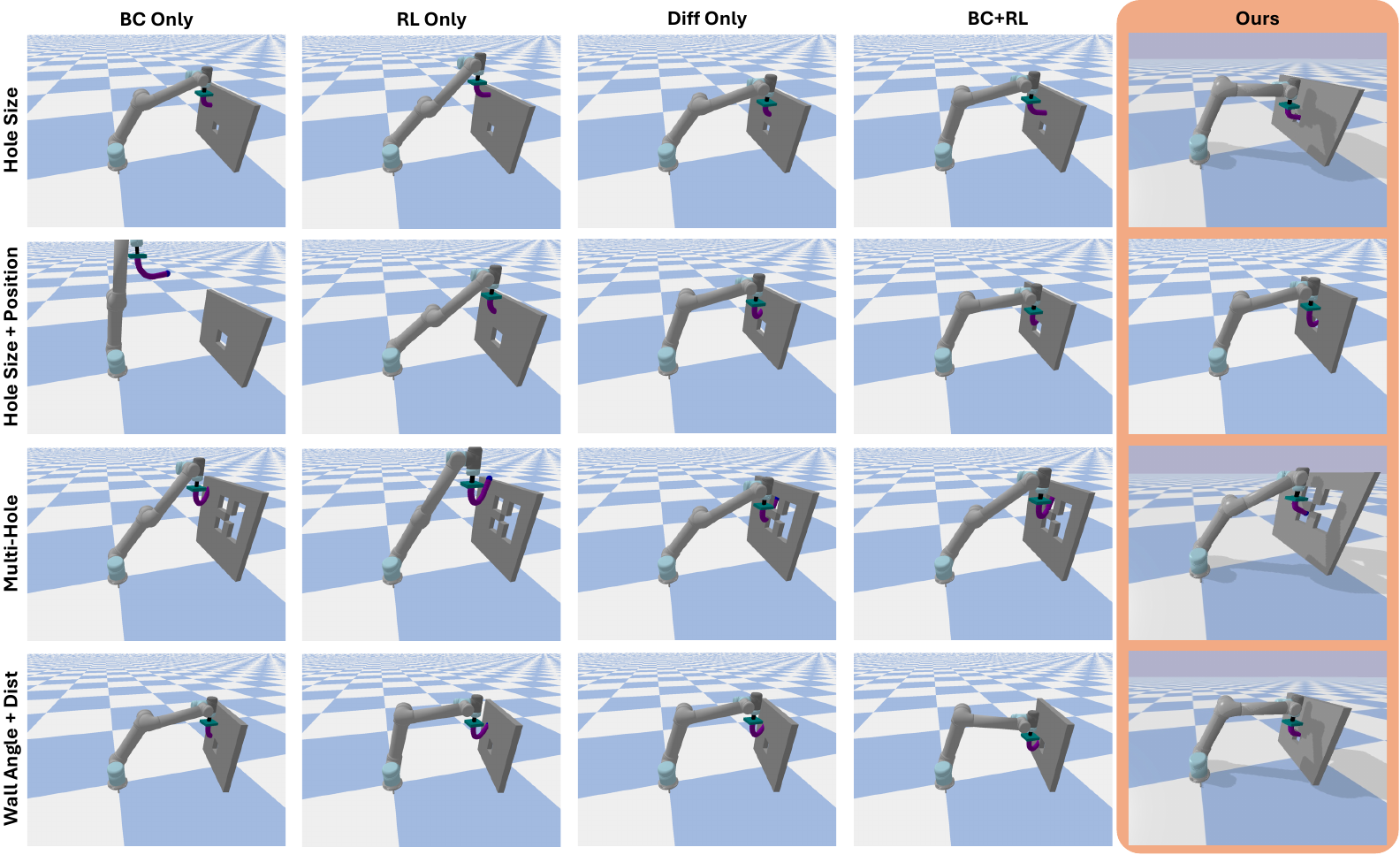}
  \caption{\textbf{Qualitative comparison across simulation environments.} Each row varies a different aspect of environment geometry: size of the aperture (Hole Size), combined hole size and position (Hole Size + Position), multiple apertures requiring global shape reasoning (Multi-Hole), and simultaneous wall angle and standoff distance variation (Wall Angle + Dist). Each column shows the final backbone configuration of a different method. THREAD (Ours) consistently threads the soft segment through the target aperture while maintaining low strain and collisions.}
  \label{fig:exp}
  \vspace{-5mm}
\end{figure*}

\noindent\textbf{Receding-horizon replanning.} Every 32 steps before threading, the diffusion planner is re-invoked with the current backbone state and goal to produce a fresh plan, resetting the plan frame index to $0$. This corrects drift between the plan and the real robot state. Replanning is suppressed after threading to prevent the tip from routing back through the wall.

\subsection{Training Details}
\noindent The diffusion denoiser is a $6$-layer Transformer with $8$ attention heads and model dimension $512$, trained with 277 expert demonstrations collected via tele-operation. The residual SAC policy is trained for 1M timesteps using Stable-Baselines3~\cite{stable-baselines3}, with episodes capped at 256 steps. A curriculum progressively reduces the hole size across five stages: $20 \to 15 \to 12 \to 10 \to 5$ cm, advancing when the rolling 50-episode success rate exceeds 30\% (with a minimum of 200 episodes per stage).

\section{Experiments and Results}
\subsection{Experimental Setup}
\begin{table}[t!]
\centering
\caption{\textbf{Overall quantitative results} Averaged across all test environments, evaluated on held-out instances varying hole size, position, and count. }
\resizebox{\linewidth}{!}{%
\begin{tabular}{l c c c c}
\toprule
Method & Success (\%) $\uparrow$ & Tip Dist (m) $\downarrow$ & Collisions (\%) $\downarrow$ & Smoothness $\downarrow$ \\
\midrule
BC Only & 26.2 & 0.104 $\pm$ 0.032 & 38.5 & 0.096 $\pm$ 0.001 \\
RL Only & 38.2 & 0.080 $\pm$ 0.011 & 61.4 & 0.110 $\pm$ 0.002 \\
Diff Only &  60.4 & 0.057 $\pm$ 0.013 & \textbf{9.3} & 0.024 $\pm$ 0.001 \\
BC + RL & 75.3 & 0.063 $\pm$ 0.020 & 62.2 & 0.076 $\pm$ 0.000 \\
\textbf{Ours} & \textbf{92.4} & \textbf{0.015 $\pm$ 0.010} &  12.2 &\textbf{0.008} $\pm$ \textbf{0.001} \\
\bottomrule
\end{tabular}}
\label{tab:overall}
\vspace{-6mm}
\end{table}

\noindent All training is conducted in SoftManiSim~\cite{kasaeisoftmanisim} simulation environment based on PyBullet. As shown in Figure \ref{fig:system_fig} (a), the robot consists of a 6-DOF UR5 rigid arm with a two-segment SCA mounted at the end-effector, giving a 9-dimensional action space (6D soft curvature (3 from each segment), 3D end-effector delta). 
The real-world experimental setup, shown in Figure \ref{fig:system_fig} (b), consists of a B3 (three bending actuators) SCA attached to a 6-DOF myCobot 280 rigid arm. The SCA is fabricated from a combination of Fiber Reinforced Elastomeric Enclosures (FREEs)~\cite{uppalapati2018towards}, which are cost effective and safe in unstructured environments. 
The SCA is actuated by three independent pneumatic channels, giving control inputs $(b_1, b_2, b_3)$ corresponding to the three bending actions. The rigid arm is controlled via its six joint angles $(j_1, j_2, j_3, j_4, j_5, j_6)$. A Polhemus magnetic sensor provides the tip pose for training data collection and online RL, and a lightweight RGB camera is mounted on the SCA tip to collect RGB images for point cloud reconstruction. 
Critically, no environment-specific retraining is performed; real-world deployment requires only $\sim$100 online RL update steps, validating the sample efficiency of THREAD's sim-to-real transfer.\looseness=-1


\noindent\textbf{Metrics.} 
We evaluate task completion, execution safety and trajectory quality with the following metrics.

\textit{Success Rate (\%)}: The percentage of trials in which the robot tip comes within 2cm of the goal position at the end of the trajectory, indicating successful task completion.

\textit{Tip Distance (m)}: The Euclidean distance between the robot's final tip position and the goal position, averaged across all trials; lower values indicate more precise reaching.

\textit{Collisions (\%)}: The percentage of trials in which any point on the robot backbone intersects the wall slab outside the hole region during execution.

\textit{Smoothness}: Calculated using Eq.~\ref{eq:smoothness_loss}, where lower values indicate smoother trajectories with fewer abrupt motion changes.

\begin{table*}[htp!]
\centering
\caption{\textbf{Quantitative results across environment configurations.} Per-environment comparison of all methods across three levels of geometric complexity: varying hole size, varying hole size and position, and varying hole size, position, and number of holes. THREAD's advantage over baselines grows with environment difficulty, most pronounced in the multi-hole setting where global shape reasoning is required.}
\resizebox{\textwidth}{!}{%
\begin{tabular}{l@{\hskip 4mm} l c c c c c c c c c c c c}
\toprule
& & \multicolumn{4}{c}{Hole Size}
  & \multicolumn{4}{c}{Hole Size + Position}
  & \multicolumn{4}{c}{Hole Size + Position + Number} \\
\cmidrule(lr){3-6} \cmidrule(lr){7-10} \cmidrule(lr){11-14}
\# & Method
& Success (\%) $\uparrow$ & Tip Dist (m) $\downarrow$ & Collisions (\%) $\downarrow$ & Smoothness $\downarrow$
& Success (\%) $\uparrow$ & Tip Dist (m) $\downarrow$ & Collisions (\%) $\downarrow$ & Smoothness $\downarrow$
& Success (\%) $\uparrow$ & Tip Dist (m) $\downarrow$ & Collisions (\%) $\downarrow$ & Smoothness $\downarrow$ \\
\midrule
1 & BC Only
& 56.6 & 0.055 $\pm$ 0.001 & 45.5  & 0.096 $\pm$ 0.008
& 16.0 & 0.081 $\pm$ 0.010  & 61.0 & 0.093 $\pm$ 0.005
& 6.0 & 0.177 $\pm$ 0.026  & 8.9 & 0.099 $\pm$ 0.002 \\
2 & RL Only
& 58.1  & 0.044 $\pm$ 0.015 & 56.6 & 0.10 $\pm$ 0.005
& 55.5& 0.077 $\pm$ 0.023 & 60.0 & 0.142 $\pm$ 0.004
& 1.0 & 0.119 $\pm$ 0.020 & 67.7  & 0.084 $\pm$ 0.004\\
3 & Diff only
& 80.2 & 0.050 $\pm$ 0.023 & \textbf{11.2} & 0.025 $\pm$ 0.001
& 55.5 & 0.058 $\pm$ 0.030 & \textbf{6.7} & 0.026 $\pm$ 0.000
& 45.5 & 0.063 $\pm$ 0.041 & \textbf{10.0} &  0.021 $\pm$ 0.001\\
4 & BC + RL
& 83.3 & 0.020 $\pm$ 0.001 & 66.6 & 0.093 $\pm$ 0.007
&81.6 & 0.072 $\pm$ 0.051  & 60.0 & 0.086 $\pm$ 0.000
& 61.0 & 0.097 $\pm$ 0.020 & 60.0 & 0.048 $\pm$ 0.009 \\
5 & \textbf{Ours}
& \textbf{95.3} & \textbf{0.010 $\pm$ 0.004} & 11.5 & \textbf{0.002 $\pm$ 0.001}
& \textbf{92.1} & \textbf{0.011 $\pm$ 0.010} & 15.0 & \textbf{0.003 $\pm$ 0.001}
& \textbf{89.9} & \textbf{0.021 $\pm$ 0.005} & \textbf{10.0} & \textbf{0.019 $\pm$ 0.009}\\
\bottomrule
\end{tabular}}
\label{tab:env_specific}
\vspace{-3mm}
\end{table*}

\begin{table*}[htp!]
\centering
\caption{\textbf{Generalization to out-of-distribution environments.} Success rate and trajectory quality metrics evaluated on two held-out environment types, not seen during training: varying wall angle and varying wall distance, while also varying hole sizes and positions.}
\resizebox{\textwidth}{!}{%
\begin{tabular}{l@{\hskip 4mm} l c c c c c c c c c c c c}
\toprule
& & \multicolumn{4}{c}{Wall Angle}
  & \multicolumn{4}{c}{Wall Distance} \\
\cmidrule(lr){3-6} \cmidrule(lr){7-10}
\# & Method
& Success (\%) $\uparrow$ & Tip Dist (m) $\downarrow$ & Collisions (\%) $\downarrow$ & Smoothness $\downarrow$
& Success (\%) $\uparrow$ & Tip Dist (m) $\downarrow$ & Collisions (\%) $\downarrow$ & Smoothness $\downarrow$\\
\midrule
1 & BC Only
& 2.52 & 0.300 $\pm$ 0.073 & \textbf{5.0} & 0.098 $\pm$ 0.015
& 1.09 & 0.111 $\pm$ 0.058 & 15.0 & 0.091 $\pm$ 0.010 \\
2 & RL
& 30.30 & 0.113 $\pm$ 0.120 & 100 & 0.101 $\pm$ 0.037
& 28.9 & 0.083 $\pm$ 0.026 & 77.6 & 0.057 $\pm$ 0.003\\
3 & Diff Only
& 45.5 & 0.107 $\pm$ 0.026 & 36.7 & 0.014 $\pm$ 0.001
& 55.5 & 0.080 $\pm$ 0.026 & 25.0&  0.018 $\pm$ 0.001 \\
4 & BC+RL
& 65.5 & 0.059 $\pm$ 0.056 & 75.0 & 0.012 $\pm$ 0.004
& 66.6 & 0.074 $\pm$ 0.034 & 66.6 & 0.103 $\pm$ 0.003  \\
5 & \textbf{Ours}
& \textbf{87.6} & \textbf{0.020 $\pm$ 0.011} & 6.7 & \textbf{0.010 $\pm$ 0.003}
& \textbf{88.8} & \textbf{0.019 $\pm$ 0.020} & \textbf{10.0} & \textbf{0.010 $\pm$ 0.003}  \\
\bottomrule
\end{tabular}}
\label{tab:wall_ang}
\vspace{-6mm}
\end{table*}

\noindent\textbf{Baselines.} We compare THREAD against four baselines that ablate its key components and span the design space of learning-based approaches to soft manipulation. All baselines use the same simulation environments, reward functions, and evaluation protocol as~\ourmethod{}, and those trained with demonstrations use the same dataset.

~\bconly~\cite{nazeer2023soft}: A visuomotor policy trained via behavioural cloning with DAgger-style correction. This replaces the diffusion prior with a deterministic, single-mode imitation policy.

~\rl~\cite{nazeer2024rl,nazeer2024comparison}: A SAC policy trained from scratch using the same reward structure as~\ourmethod{}, but without demonstrations or a learned prior, establishing the performance ceiling of pure RL exploration and isolating the contribution of the diffusion backbone as a shaping prior over feasible configurations.\looseness=-1

\begin{figure}[t!]
    \centering
    \includegraphics[width=0.49\linewidth]{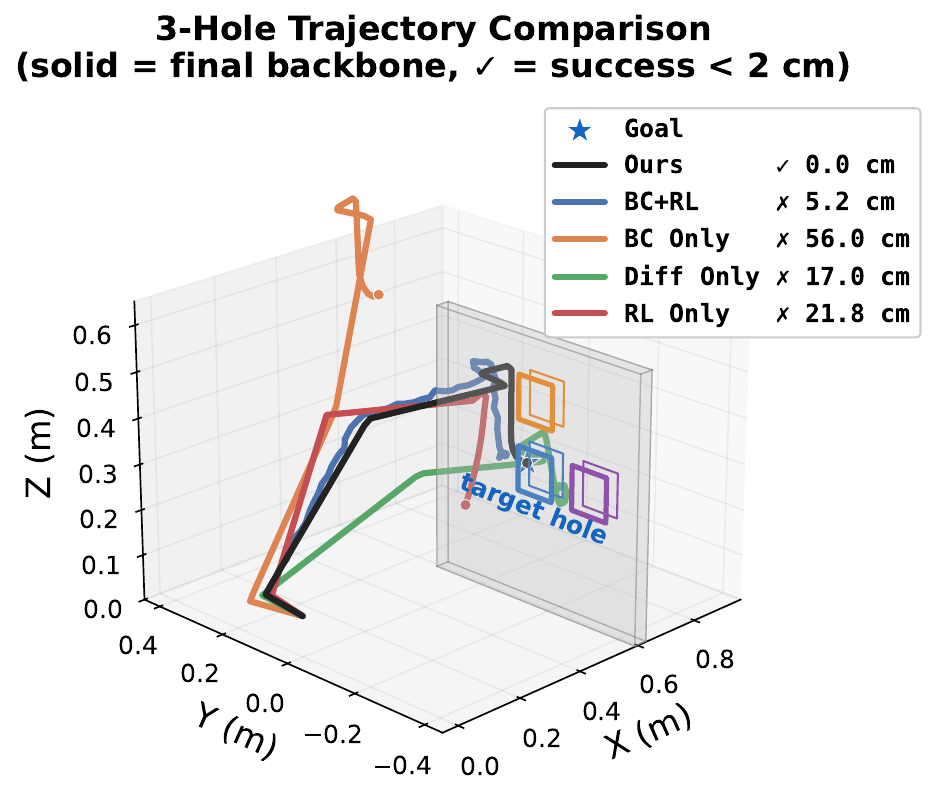}
    \includegraphics[width=0.49\linewidth]{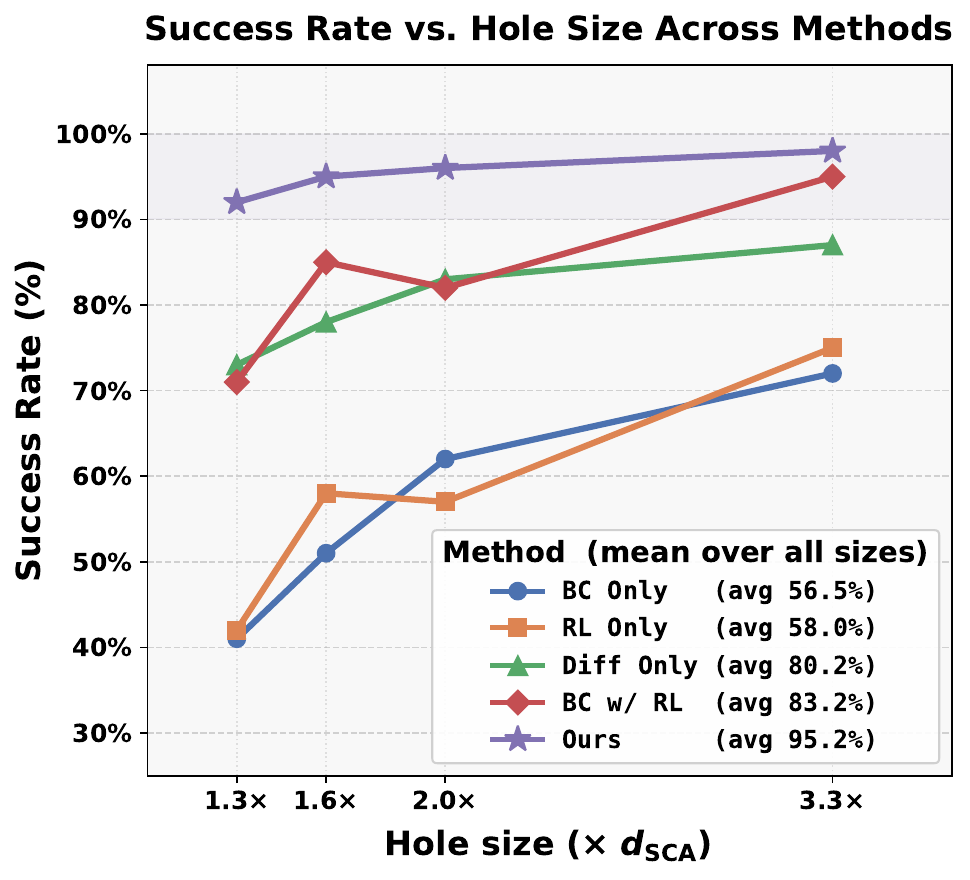}
    \caption{\textbf{Performance comparison across environments and hole sizes.} (a) Qualitative trajectories generated by each method in a multi-hole environment. (b) Success rate as a function of hole size, expressed as a multiple of the SCA diameter.~\ourmethod{} maintains higher success rates as hole size decreases, with a more gradual performance degradation compared to all baselines, demonstrating the advantage of a learned generative prior over feasible backbone shapes in highly constrained settings.}
    \label{fig:graphs}
    \vspace{-7mm}
\end{figure}

~\rlbc~\cite{nazeer2024comparison}: RL initialized from a behaviorally cloned policy via value-conditioned initialization. The strongest non-diffusion baseline, combining demonstration data with online RL refinement, but lacking multi-modal trajectory generation.

~\diff{}: The THREAD diffusion model with inference-time geometric guidance but without online RL updates or residual correction. This baseline tests whether the diffusion prior alone is sufficient, or whether closed-loop residual RL correction and MPC-style replanning are necessary.

\noindent\textbf{Testing environments.} We evaluate across five environment types of increasing geometric complexity. Training covers three configurations: varying (1) hole size, (2) hole size and position, and (3) hole size, position, and count. At test time, models are evaluated on unseen instances of these training distributions as well as two fully held-out environment types, varying wall angle and varying wall distance (while also varying hole size and position), that introduce structural variation entirely absent from training. Each test begins with the hybrid arm in a random start pose.

\subsection{Experimental Results}

\noindent\textbf{Result 1 - THREAD achieves consistent success with significantly lower collision cost.} 
Across all evaluation environments,~\ourmethod{} achieves the highest success rate (92.4\%) while minimizing tip error (0.015m) and collision rate (12.2\%) -- a $17.1\%$ improvement in success over the strongest baseline (\rlbc{}) and a $5\times$ reduction in collisions (Table~\ref{tab:overall}, Figure~\ref{fig:exp}). \rlbc{} reaches moderate task completion but at the cost of collisions exceeding 60\% across all environments, revealing that reactive RL correction without a geometric prior leads to a fundamental success-safety tradeoff that THREAD does not exhibit.~\rl{} similarly suffers 61.4\% collisions and only 38.2\% success -- worse on both dimensions. 
Notably,~\bconly{} reports comparatively low collisions, but this reflects task failure rather than geometric awareness, with the policy failing to reach the wall entirely. 
These results directly validate the closed-loop architecture: the diffusion prior provides collision-aware geometric structure that RL alone cannot discover, while the residual SAC policy ensures reliable execution that the diffusion prior alone cannot guarantee (\diff{}, Result 3). This advantage is most pronounced under extreme geometric constraints -- as shown in Table~\ref{tab:env_specific} and Fig.~\ref{fig:graphs}(b), \ourmethod{} degrades more gradually as constraints tighten. 


\begin{figure*}[htp!]
\centering
\smallskip
  \includegraphics[width=0.80\linewidth]{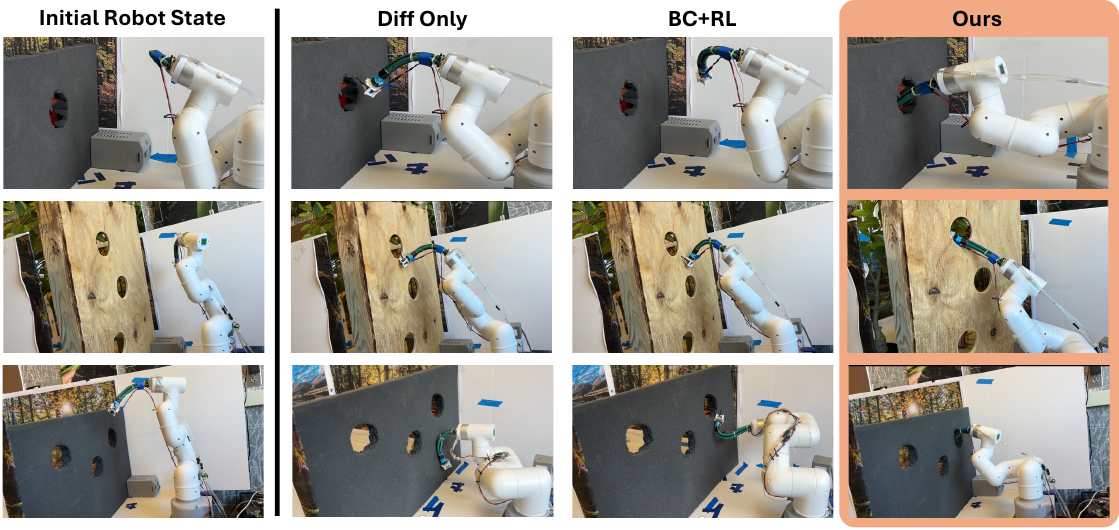}
  \caption{\textbf{Real-world deployment across environment configurations.} Initial and final states for Diffusion-only, RL+BC and THREAD across three environments with varying hole size, position, number of holes, and wall distance. THREAD successfully threads in all three configurations while baselines fail.}
  \label{fig:real-qual}
  \vspace{-5mm}
\end{figure*}

\begin{table}[t!]
\centering
\caption{\textbf{Cross-embodiment real-world transfer.} THREAD is trained in simulation on a UR5 with a two-segment SCA and deployed with only $\sim$ 100 online RL update steps and no retraining on an Elephant Robotics myCobot with a single-segment SCA. Metrics are reported across 12 trials (on the executed trajectory) per method.}
\resizebox{\linewidth}{!}{%
\begin{tabular}{l c c c c}
\toprule
Method & Success (\%) $\uparrow$ & Tip Dist (m) $\downarrow$ & Collisions (\%) $\downarrow$ \\
\midrule
Diff Only & 34.2  & 0.098 $\pm$ 0.021 & 16.6  \\
BC + RL & 41.6  & 0.052 $\pm$ 0.032  & 67.0 \\
Ours & \textbf{75.0}  & \textbf{0.015 $\pm$ 0.010} & \textbf{25.1} \\
\bottomrule
\end{tabular}}
\label{tab:real-world}
\vspace{-6mm}
\end{table}

\noindent\textbf{Result 2 - The backbone-centric formulation generalizes to out-of-distribution environments.} On environments with entirely unseen distributions (wall angle and wall distance, Table~\ref{tab:wall_ang}, last row of Fig.~\ref{fig:exp}), \ourmethod{} maintains high success rates of 87.6\% and 88.8\% respectively, while all baselines degrade substantially --~\bconly{} near-completely fails (1-2\%),~\rl{} achieves only 28-30\%, and \diff{} reaches 45-55\%. The gap between methods widens with distribution shifts, revealing that baselines overfit to specific training geometries rather than learning generalizable shape-level reasoning. The backbone representation is the structural cause: 
by planning in the space of full physical shapes rather than memorized actuator trajectories, \ourmethod{} transfers across geometric variations that break coordinate-space planners.

\noindent\textbf{Result 3 - Motion quality from diffusion backbone.}
~\ourmethod{} achieves a trajectory smoothness of 0.0082, an 11$\times$ improvement over~\rl{} (0.1108) and 9$\times$ over~\rlbc{} (0.0761). Crucially,~\diff{} achieves comparable smoothness to~\ourmethod{} despite lower success rates, confirming that the diffusion backbone, not the RL component, is responsible for smooth coordinated motion. The residual SAC policy improves task success without sacrificing this smoothness, demonstrating that the two components are complementary. For applications where both task completion and physical safety matter, such as threading through narrow apertures near sensitive structures, this quality-success combination distinguishes~\ourmethod{} from all baselines. This is visually apparent in the multi-hole setting (Fig.~\ref{fig:graphs}(a)): while baselines produce erratic or infeasible backbone configurations that fail to route through the correct aperture, \ourmethod{} generates globally coherent trajectories that thread the soft segment through the target hole while maintaining low strain across the full rigid-soft chain.\looseness=-1

\noindent\textbf{Result 4 - Sim2Real cross-embodiment transfer to unseen hardware with minimal adaptation.} 
To evaluate transfer across morphologies,
we deploy \ourmethod{} on hardware it was never trained on: an Elephant Robotics myCobot paired with a single-segment SCA, distinct from the UR5 and two-segment SCA used in all simulation training (Table~\ref{tab:real-world}, Figure~\ref{fig:real-qual}). With only $\sim$ 100 online RL update steps, requiring no demonstration data and no environment-specific retraining, \ourmethod{} achieves 75.0\% success on real hardware, compared to 41.6\% for~\rlbc{} and 34.2\% for~\diff{},  successfully threading apertures as small as 1.3$\times$ the SCA diameter.
The collision rate (25.1\%) is less than half that of \rlbc{} (67.0\%), demonstrating that the safety advantage transfers reliably to the real-world. 
This result provides evidence that a strong geometric prior in the backbone shape space along with online RL adaptation allows for cross-embodiment transfer -- a capability that neither a prior-only model (\diff{}) nor a weaker prior with RL (\rlbc{}) can achieve.

\begin{figure}
    \centering
    \includegraphics[width=\linewidth]{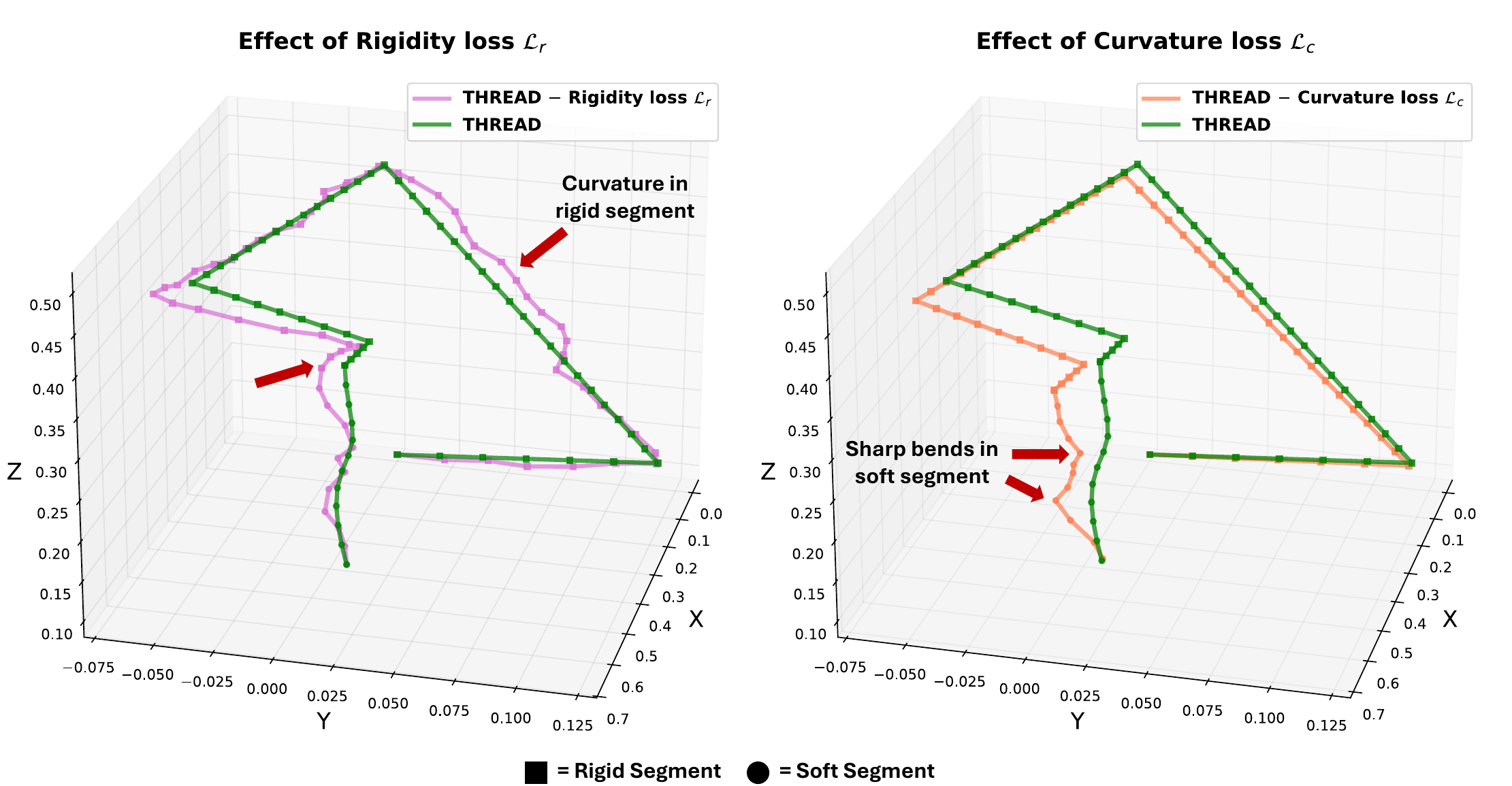}
    \vspace{-5mm}
    \caption{\textbf{Effect of rigidity and curvature diffusion losses.} Without $\mathcal{L}_r$ (left), the rigid segment curves unnaturally; without $\mathcal{L}_c$ (right), the soft segment exhibits non-smooth, physically unrealizable bends.}
    \label{fig:ablation}
    \vspace{-4mm}
\end{figure}

\noindent\textbf{Ablation on diffusion framework.}
To evaluate the design of our diffusion framework, we ablate each proposed component by comparing our full model against variants with individual components removed (Table~\ref{tab:ablation}). Inference-time SDF guidance provides the largest performance gain, with its removal resulting in $\sim 5.5\times$ more episodes ending in collision and $\sim 3.5\times$ lower success rate. Removing the temporal smoothness loss $\mathcal{L}_t$ degrades trajectory smoothness, producing abrupt inter-frame motion that reduces task success. As shown in Figure~\ref{fig:ablation}, removing the rigidity loss $\mathcal{L}_r$ introduces curvature in the rigid segment of the robot, yielding unrealizable backbone shapes, while removing the curvature loss $\mathcal{L}_c$ results in non-smooth, physically unrealizable bends in the soft segment.

\begin{table}[t!]
\centering
\caption{\textbf{Ablation on diffusion design components of THREAD.} Each row removes a single component while keeping all others fixed, evaluated across all test environments.
}
\resizebox{\linewidth}{!}{%
\begin{tabular}{l c c c c}
\toprule
Method & Success (\%) $\uparrow$ & Tip Dist (m) $\downarrow$ & Collisions (\%) $\downarrow$ & Smoothness $\downarrow$\\
\midrule
THREAD $- \space \text{Guidance}$  & 26.6  & 0.066 $\pm$ 0.031 & 66.6 & 0.009 $\pm$ 0.003\\
THREAD $-$ Temporal loss $\mathcal{L}_t$ &  77.6 & 0.048 $\pm$ 0.041   & 13.3 & 0.012 $\pm$ 0.005\\
THREAD $-$ Rigidity loss $\mathcal{L}_r$ & 85.5 & 0.037 $\pm$ 0.020& 15.0 & 0.011 $\pm$ 0.004\\
THREAD $-$ Curvature loss $\mathcal{L}_c$ & 90.0 & 0.019 $\pm$ 0.009 & 14.5 & 0.009 $\pm$ 0.007\\
THREAD & \textbf{92.4} & \textbf{0.015 $\pm$ 0.010} &  \textbf{12.2} & 0.008 $\pm$ 0.001\\
\bottomrule
\end{tabular}}
\label{tab:ablation}
\vspace{-7mm}
\end{table}

\section{Limitations and Future Work}
\noindent While our framework demonstrates the effectiveness of a diffusion-based trajectory planner for hybrid manipulators in both simulation and real-world settings, several limitations point to future directions. First, although the backbone representation reduces morphological coupling, the small online RL adaptation budget may still be prohibitive in safety-critical settings; incorporating uncertainty quantification into the planning loop could reduce this reliance and move toward zero-shot transfer. Second, our evaluations focus on planar wall threading, and scaling to sequential apertures, tunnels, and more complex environments encountered in tasks such as disaster response or surgical assistance remains an exciting direction. Finally, pairing longer SCAs with higher-DOF or mobile rigid bases would expand the operational workspace and enable more challenging navigation tasks.


\section{Conclusion}
\noindent We presented THREAD, the first diffusion-based trajectory planner for hybrid rigid-soft manipulation through constrained apertures. By planning in continuous backbone shape space rather than tip or joint space, THREAD learns a generative prior over physically realizable threading solutions that captures the multi-modal feasibility structure that reactive and imitation-based methods cannot represent. Beyond task performance, the backbone-centric representation demonstrates a structural property that purely reactive planners cannot offer: the learned geometric prior transfers across morphologically distinct hardware without retraining, suggesting a path toward general-purpose planners for the broad class of soft and hybrid manipulators that share backbone shape as a common representational primitive.

\addtolength{\textheight}{-12cm}   




\bibliographystyle{IEEEtran}{}
\bibliography{references}

\end{document}